\documentclass[lettersize,journal]{IEEEtran}
\usepackage{amsmath,amssymb,amsfonts}
\usepackage{algorithmic}
\usepackage{array}
\usepackage[caption=false,font=normalsize,labelfont=sf,textfont=sf]{subfig}
\usepackage{textcomp}
\usepackage{stfloats}
\usepackage{url}
\usepackage{verbatim}
\usepackage{graphicx}
\usepackage[dvipsnames]{xcolor}
\def\BibTeX{{\rm B\kern-.05em{\sc i\kern-.025em b}\kern-.08em
    T\kern-.1667em\lower.7ex\hbox{E}\kern-.125emX}}
\usepackage{balance}
\usepackage{cite}
\usepackage{multirow, diagbox, makecell, colortbl} 

\begin{document}
\title{FusionLoc: Camera-2D LiDAR Fusion Using Multi-Head Self-Attention for End-to-End Serving Robot Relocalization}
\author{Jieun Lee, Hakjun Lee, and Jiyong Oh*
\thanks{J. Lee and J. Oh are with Robot IT Convergence Research Section, Daegu-Gyeongbuk Research Center, Electronics and Telecommunications Research Institute (ETRI), Daegu, Korea (e-mail: \{jieun.lee, jiyongoh\}@etri.re.kr). H. Lee is with Polaris3D, Pohang, Korea (e-mail: hakjunlee@polaris3d.co). J. Oh is the corresponding author.
}}



\maketitle

\begin{abstract}
As technology advances in autonomous mobile robots, mobile service robots have been actively used more and more for various purposes.
Especially, serving robots have been not surprising products anymore since the COVID-19 pandemic.
One of the practical problems in operating a serving robot is that it often fails to estimate its pose on a map that it moves around.
Whenever the failure happens, servers should bring the serving robot to its initial location and reboot it manually.
In this paper, we focus on end-to-end relocalization of serving robots to address the problem.
It is to predict robot pose directly from only the onboard sensor data using neural networks.
In particular, we propose a deep neural network architecture for the relocalization based on camera-2D LiDAR sensor fusion.
We call the proposed method FusionLoc.
In the proposed method, the multi-head self-attention complements different types of information captured by the two sensors to regress the robot pose.
Our experiments on a dataset collected by a commercial serving robot demonstrate that FusionLoc can provide better performances than previous end-to-end relocalization methods taking only a single image or a 2D LiDAR point cloud as well as a straightforward fusion method concatenating their features.
\end{abstract}


\section{Introduction}
\label{sec:introduction}
Recently, various mobile robots can be seen in many places around us for reasons such as the development of autonomous mobile robot technology, the pursuit of efficiency in repetitive tasks, and the increase in the value of non-face-to-face services. For instance, some mobile robots are helping people in various places, such as guiding people at airports or museums, disinfecting schools or hospitals, following people with heavy objects, and serving food in restaurants.
Focusing on in restaurants and cafes, servers are known to walk around 8 to 15 kilometers a day. Also, carrying dishes in hand is more laborious than walking empty-handed. As a result, it can lower the quality of essential services for customers. For these reasons, serving robots are employed in more and more sites to improve more efficient working environments while reducing physical burden primarily.

For the autonomous mobile robots, localization is indispensable.
Some mobile robots use a global positioning solution based on markers attached to the ceiling for localization.
However, it has limitations due to the cost of infrastructure construction and maintenance and the damage to aesthetic elements in cafes and restaurants.
Another localization methodology first produces a map of the place where the robot operates using simultaneous localization and mapping (SLAM) methods \cite{thrun2005}, \cite{Hong2021}. Then, the serving robot estimates its location on the map based on information captured from its various sensors like wheel encoders, cameras, LiDARs, or IMUs. However, it often fails to estimate its location for some reasons, e.g., many people around the robot. The failure in the localization requires the serving robot user to reboot it at a predetermined starting position and orientation. That is one of the reasons why relocalization is needed.

\begin{figure*}[t]
\begin{center}
    \includegraphics[width=17cm]{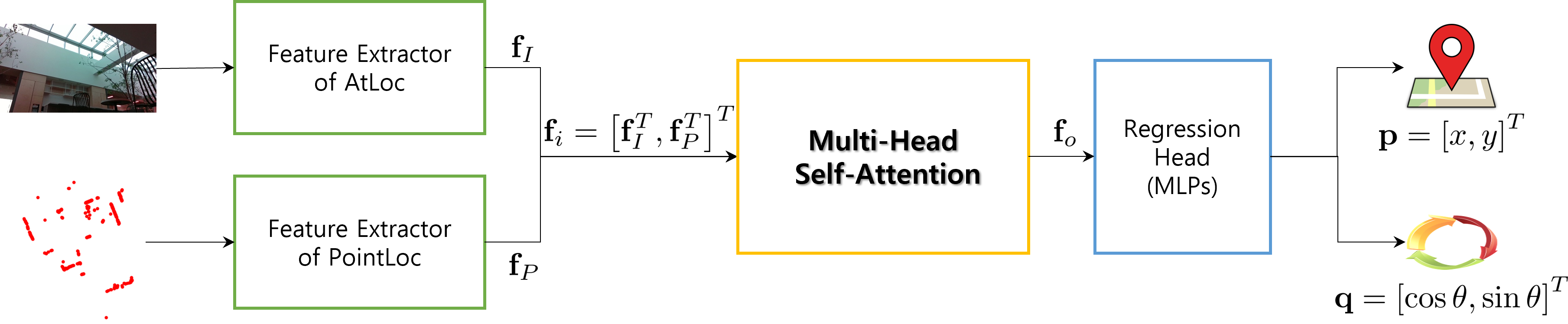}
    \caption{The overall structure of the proposed method for serving robot relocalization based on camera-2D LiDAR fusion. It takes an image and a point cloud as input, and outputs the robot position $\mathbf{p}$ and orientation $\mathbf{q}$.}
\label{fig:overall-structure}
\end{center}
\end{figure*}

For mobile robot relocalization, many studies were presented based on conventional visual features, and some studies utilized visual information together with geometric information measured by a LiDAR sensor \cite{Su2017}.
However, deep learning studies for relocalization have been actively conducted in the recent few years.
PoseNet \cite{Kendall2015} applied a convolutional neural network (CNN) to a single RGB image to regress a 6 degrees of freedom (DoF) camera pose. It was shown that PoseNet is more robust to illumination variation and occlusion than point feature-based relocalization methods.
This advantage is one of the reasons why we focus on deep learning-based relocalization methods.
Various studies followed PoseNet to apply a Bayesian CNN \cite{Kendall2016}, long-short term memory (LSTM) \cite{Walch2017}, or geometric reprojection loss function \cite{Kendall2017}.
Also, temporal information was utilized with an LSTM module \cite{Ronald2017}, and an encoder-decoder architecture was employed for camera relocalization in \cite{Melekhov2017}. Moreover, geometric constraints of two consecutive poses were considered to improve relocalization accuracy in \cite{Brahmbhatt2018}, and a self-attention block was adopted in regressing the camera pose \cite{Wang2020}. In recent studies, deep neural networks (DNN) took LiDAR \cite{Wang2022} or IMU \cite{Lang2019} data instead of using images for relocalization in an end-to-end manner.

In this paper, we leverage two sensors equipped with a commercial serving robot to improve relocalization accuracy. From the perspective of the extension of previous studies, this paper proposes a fusion DNN taking both an RGB image and a 2D LiDAR point cloud as input to regress the 3-DoF pose of the serving robot. We call the proposed architecture \emph{FusionLoc}.
To our best knowledge, this work is the first deep learning-based study fusing camera and LiDAR to address the mobile robot relocalization in two-dimensional planar space.
The proposed network extracts features by adopting the AtLoc \cite{Wang2020} architecture and the PointLoc \cite{Wang2022} architecture from RGB images and 2D LiDAR point clouds, respectively.
These features are combined through a concatenation operation. Then, the information captured from each sensor interacts together within multi-head self-attention (MHSA) layers \cite{Vaswani2017}.
Finally, FusionLoc outputs the position and orientation of the serving robot. Furthermore, this study introduces a new dataset consisting of tuples of an RGB image, a 2D LiDAR point cloud, and a 3-DoF pose. The data were collected using a Polaris3D serving robot named ereon.
Experiments on the dataset show that the proposed network outperforms the previous methods taking a single image or a 2D LiDAR point cloud only.
The contributions of this paper are summarized as follows:
\begin{itemize}
    \item A fusion deep neural network leveraging multi-head self-attention layers is proposed to take an image and a 2D LiDAR point cloud as input.
    \item A new dataset, collected using a commercial serving robot, is introduced.
\end{itemize}

This paper is organized as follows.
Section \ref{sec:relatedworks} briefly introduces the previous DNN-based studies for relocalization.
Section \ref{sec:method} describes the proposed DNN architecture for camera-2D LiDAR fusion in detail.
The new dataset is explained, and the experiments based on the dataset are shown in Section \ref{sec:experiments}. Finally, Section \ref{sec:conclusions} concludes the paper and provides future work.

\section{Related works}
\label{sec:relatedworks}

\subsection{Deep learning-based camera relocalizaton}
There have been a lot of studies on deep learning for localization using cameras.
The visual localization studies include place recognition and end-to-end camera relocalization.
Given a query image, place recognition performs localization by image retrieval, i.e., transforming the query image into a descriptor by a deep network and then searching the most similar descriptor with the input descriptor in a database consisting of descriptors and their location information.
NetVLAD \cite{Arandjelovic2016} is a representative study belonging to visual place recognition.
On the other hand, end-to-end camera relocalization directly predicts the pose of the mobile robot from the input image.
PoseNet \cite{Kendall2015} was a breaking ground work that tried to directly regress 6-DoF camera pose from a single image using a CNN.
It was noted that it was robust to motion blur, darkness, and unknown camera intrinsics compared to conventional methods based on the scale-invariant feature transform (SIFT \cite{Lowe2004}).
Probabilistic PoseNet \cite{Kendall2016}, an extension of PoseNet, adopted a Bayesian CNN with Monte Carlo dropout sampling to handle uncertainty in pose estimation.
In \cite{Walch2017}, LSTM was presented as a substitute for the fully-connected layer before the pose regression layer to prevent overfitting and to perform a structured dimensionality reduction.
In \cite{Kendall2017}, loss function was also a consideration to improve relocalization performance.
To this end, two alternatives were presented and tested in the study.
One was a learnable balancing parameter between position and orientation, and the other was the reprojection error between predicted and actual camera poses.
In \cite{Ronald2017}, VidLoc was proposed.
It leveraged a bidirectional LSTM to utilize temporal information in successive images.
The trial led to some reduction in the relocalization error.
In \cite{Melekhov2017}, the authors employed an encoder-decoder CNN architecture, which was called hourglass, for fine-grained information restoration.
In the encoder, ResNet-34 \cite{He2016} was adopted instead of GoogLeNet \cite{Szegedy2015} used in PoseNet.
In \cite{Brahmbhatt2018}, MapNet was proposed using an additional loss term related to the relative poses between image pairs together with the loss term of the absolute poses of images.
It was meaningful concerning encoding geometric constraints between consecutive poses into the loss function.
Moreover, in the study, the logarithm of the unit quaternion was also proposed as the representation of the camera orientation instead of the unit quaternion.
The logarithm of the unit quaternion has been popular in most follow-up studies.
In \cite{Laskar2017} and \cite{Balntas2018}, a Siamese architecture was presented to reduce the relocalization error with the relative poses of the image pairs.
ViPNet \cite{Hu2020} utilized the squeeze-and-excitation blocks and a Bayesian CNN with ResNet-50 to deal with uncertainty in predicting a camera pose like Probabilistic PoseNet \cite{Kendall2016}.
AtLoc \cite{Wang2020} adopted a self-attention module to focus on geometrically rigid objects rather than dynamic ones.
VMLoc \cite{Zhou2021} encoded the features of depth images (or projection of sparse LiDARs to RGB images) and RGB images in each CNN branch.
Then, the two types of features were fused by Product-of-Experts.

\subsection{Deep learning-based other sensors relocalizaton}
There have been a few studies on deep learning for relocalization with other sensors.
Compared to camera relocalization, those studies have recently been published.
The 3D LiDAR point clouds have more geometric information because they can see with a wide angle of 360 degrees and vertical field of view.
In \cite{Wang2019}, a point cloud odometry method was proposed, which took two consecutive 3D point clouds as input and predicted the transformation between them.
Each point cloud was first encoded to a panoramic depth image, and then the two depth images were stacked.
The DeepPCO network estimated 3D translation and 3D orientation from the stacked depth image using its two sub-networks.
Different from \cite{Wang2019}, PointLoc \cite{Wang2022} regressed a 6-DoF pose of a 3D LiDAR sensor directly from a point cloud.
In PointLoc, the PointNet++ \cite{Qi2017NIPS} architecture was employed to extract features from unordered and unstructured 3D point clouds, and a self-attention module was also employed to remove outliers.
Unlike DeepPCO and PointLoc, StickyLocalization \cite{Fischer2022} relocalized input 3D point cloud within a pre-built map.
In the pillar layer where PointPillar \cite{Lang2019} method was utilized, local and global key points were extracted from a current point cloud and the global point cloud corresponding to a map, respectively.
Then, self-attention and cross-attention modules in the graph neural network layer aggregated context information to improve robustness.
The final optimal transport layer output the pose of the current point cloud by a matching process between the outputs of the graph neural network layer.
Although its task was not to estimate the pose of a mobile robot, 2DLaserNet \cite{Kaleci2022} processed 2D laser scan data with a neural network developed from PointNet++ \cite{Qi2017NIPS} to classify the location of the mobile robot as one of room, corridor, and doorway.

In addition to camera and LiDAR, IMU has also been used for relocalization. 
RoNIN \cite{Herath2020} utilized the backbone networks such as ResNet, LSTM, and temporal convolutional network to estimate human 3-DoF poses from a sequence of IMU sensor data.
IDOL \cite{Sun2021} regressed 5-DoF poses with a two-stage procedure consisting of an orientation module using extended Kalman filters (EKFs) and LSTMs and a position module using bidirectional LSTMs.
In NILoc \cite{Herath2022}, a neural inertial navigation technique was presented to convert IMU sensor data to a sequence of velocity vectors.
The methods leveraged a Transformer-based neural network architecture to reduce high uncertainty in IMU data.
Moreover, in \cite{Li2021}, a neural network framework was proposed to handle laser scan data and IMU sensor data together for mobile robot localization.
For feature extraction in the method, a stack of two laser scans and a IMU data sequence between the two laser scans were passed through CNN and LSTM, respectively.
Then, another LSTM regressed the 3-DoF robot pose from the fused feature.

\begin{figure*}[t]
\begin{center}
    \includegraphics[width=17cm]{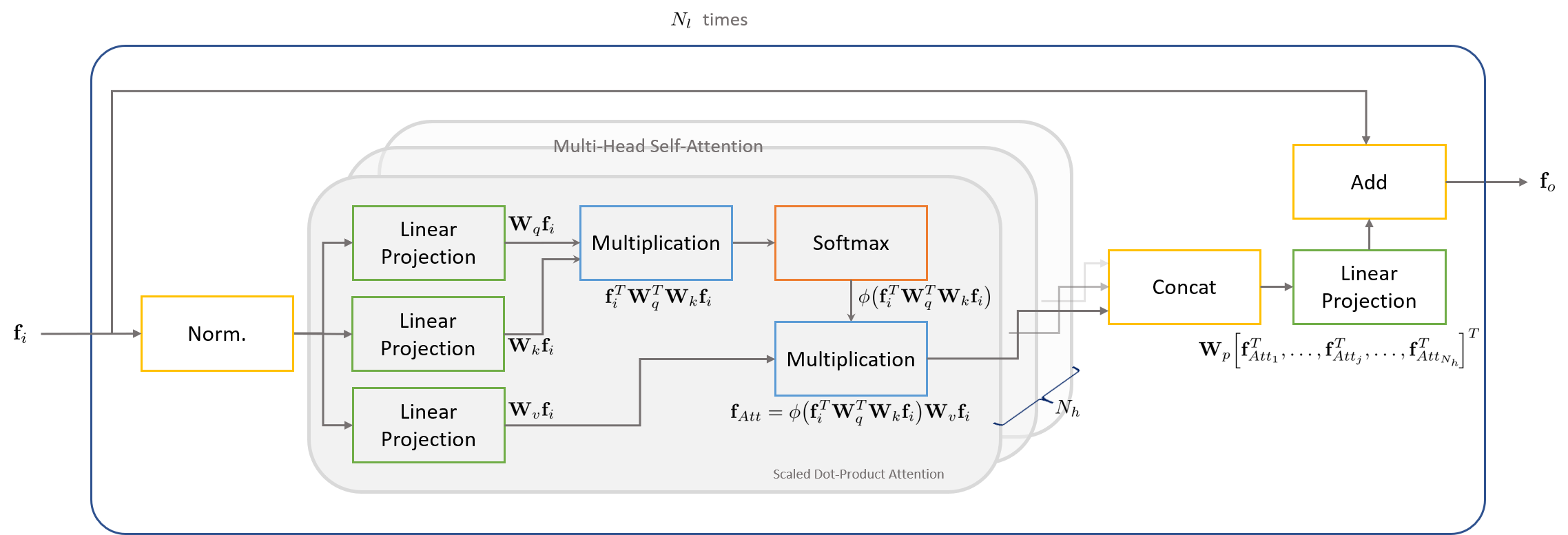}
    \caption{The procedure in the multi-head self-attention module \cite{Wang2020}.}
\label{fig:MHSA}
\end{center}
\end{figure*}

\section{Method} \label{sec:method}

\subsection{Self-attention} \label{subsec:self-attention}
Since Transformer \cite{Vaswani2017} made a great success in the literature on natural language process, its self-attention, one of the essential elements in Transformer, has been utilized in many studies for computer vision applications.
In particular, it was shown in \cite{Wang2020} that a self-attention on CNN features was effective in camera pose regression.
We use the same self-attention, but its input consists of image features computed from an input image and point cloud features computed from an input 2D point cloud.
The input of the self-attention $\mathbf{f}_{i}$, which is a column vector, is first projected to generate query $\mathbf{W}_{q}\mathbf{f}_{i}$, key $\mathbf{W}_{k}\mathbf{f}_{i}$, and value $\mathbf{W}_{v}\mathbf{f}_{i}$ by three learnable projections.
Then, the value is weighted based on the normalized correlations between the query and the key.
The correlations are calculated using the softmax function $\phi$, and these procedures can be represented as
\begin{equation} \nonumber
    \mathbf{f}_{Att} = \phi \! \left( \mathbf{f}_{i}^{T} \mathbf{W}_{q}^{T} \mathbf{W}_{k} \mathbf{f}_{i} \right)\! \mathbf{W}_{v} \mathbf{f}_{i}.
\label{equ:attention1}
\end{equation}
Here, $\mathbf{f}_{Att}$ is the output of scaled dot-product attention in \cite{Vaswani2017}. 
The output of the self-attention $\mathbf{f}_{o}$ is computed based on a linear projection of $\mathbf{f}_{Att}$ and a residual connection as
\begin{equation} \nonumber
    \mathbf{f}_{o} = \mathbf{W}_{p} \mathbf{f}_{Att} + \mathbf{f}_{i}.
\label{equ:attention2}
\end{equation}
According to \cite{Wang2020}, it was illustrated that this self-attention makes activation more intense in fixed objects like buildings and furniture in its input image than in dynamic ones like moving vehicles and pedestrians, which leads to robust relocalization.
It was also shown that the self-attention is effective in capturing the correlations among the elements of $\mathbf{f}_{i}$ in camera relocalization as well as in other computer vision applications \cite{Fu2019}, \cite{Yu2020}.

\subsection{Camera-LiDAR fusion for relocalization with multi-head self-attention} \label{subsec:fusion}
The proposed network consists of two feature extraction modules, an MHSA module, and a regression module as shown in Fig. \ref{fig:overall-structure}.
Receiving two types of data obtained from different sensors, an image captured by a camera and a point cloud captured by a 2D LiDAR, it provides a robot pose, which consists of a position $\mathbf{p}$ and an orientation $\mathbf{q}$.
We assume that they are synchronized in time.

At first, in the feature extraction modules, the image and point cloud features are computed from the image and the point cloud, respectively.
The image feature $\mathbf{f}_{I}$ is computed by the feature extractor of AtLoc \cite{Wang2020}, which consists of a CNN based on ResNet-34 \cite{He2016} followed by the self-attention module described above.
On the other hand, the point cloud feature $\mathbf{f}_{P}$ is computed by the feature extractor of the PointLoc \cite{Wang2022}, which consists of the set abstraction layers, the self-attention layer, and the group all layer.
Except for the self-attention layers, the other layers in the network were presented in PointNet++ \cite{Qi2017NIPS}, which was proposed for 3D point cloud classification and semantic segmentation.
Since the input point cloud is two-dimensional, not three-dimensional, we modified the feature extraction module so that it receives a 2D point cloud as input.
As in a CNN where a basic block consisting of convolution, nonlinear activation, and pooling operations is consecutively performed, several successive set abstraction layers extract features from a 2D LiDAR scan to give a feature matrix.
The self-attention layer in PointLoc is slightly different from the one in AtLoc described above.
In the layer, a weight matrix is computed by passing the feature matrix through a shared multi-layer perceptron (MLP) layer, a sigmoid function, and a broadcasting operation.
Then, the output of the self-attention layer is obtained by multiplying the feature matrix and the weight matrix element-wise.
Note that the input and output of the self-attention layer in PointLoc also have the same dimension as in AtLoc.
The group all layer takes the output of the self-attention layer as its input and provides the point cloud feature $\mathbf{f}_{P}$ by conducting an MLP and max-pooling operations.
More details are referred to as \cite{Wang2022}.

From the perspective of sensor fusion, it may be important to make data measured from different sensors interact with each other.
For the relocalization using both camera and 2D LiDAR, the most straightforward way to use their data together in a network is to combine the two features by summation or concatenation as in \cite{Oertel2020}, \cite{Pan2020}, \cite{Komorowski2021} and perform a pose regression using the combined feature, which is called the fusion feature below.
In our network, we chose concatenation to make the fusion feature from the image and point cloud features because summation is impossible but concatenation allows them to have different dimensions.
However, we figured out from many experiments that those simple concatenation is not enough to effectively fuse different information obtained from the two sensors.
To overcome this limitation, we propose to apply additional MHSA as shown in Fig. \ref{fig:MHSA} to the fusion feature. 
Since the self-attention effectively captures the correlations between its input elements, it can allow different information contained in $\mathbf{f}_{I}$ and $\mathbf{f}_{P}$ to interact with each other.
In other words, we utilize the MHSA module for information fusion.

In \cite{Vaswani2017}, MHSA was presented to capture different correlations among input elements by performing several scaled dot-product attentions in parallel.
To this end, the outputs of all attentions are integrated and a linear projection is performed as
\begin{equation} \nonumber
    \mathbf{W}_{p} \! \left[ \mathbf{f}_{Att_{1}}^{T}, \ldots, \mathbf{f}_{Att_{j}}^{T}, \ldots, \mathbf{f}_{Att_{N_{h}}}^{T} \right]^{T},
\label{equ:multi_attention}
\end{equation}
where $\mathbf{f}_{Att_{j}}$ is the output of the $j$-th scaled dot-product attention, $N_{h}$ is the number of the attention heads, and $\left[ \mathbf{a}_{1}^{T}, \ldots, \mathbf{a}_{n}^{T} \right]$ is the concatenation of $\{ \mathbf{a}_{i}^{T} \}_{i=1}^{n}$.
In this operation, each attention is scaled by a scaling factor $N_{h}$ so that its output has the same dimension as the input.
Like the Transformer encoder in \cite{Vaswani2017}, a normalization layer is applied before MHSA, and a residual connection is attached after MHSA.
We employ batch normalization (BN, \cite{Ioffe2015}) instead of layer normalization (LN, \cite{Ba2016}) different from \cite{Vaswani2017}.
It was demonstrated in \cite{Ioffe2015} that LN is more effective than BN for recurrent networks.
However, we find out from experiments that BN is more effective than LN in this work.
Also, we do not use the positional encoding, another input of the Transformer encoder, because the order of elements in the fusion feature is not important in this task, unlike a sequence.
This MHSA block with identical architecture repeats $N_{l}$ times as in \cite{Vaswani2017}.
It will be demonstrated in experiments that the repetition of MHSA improves the accuracy of the relocalization based on the camera-2D LiDAR fusion.

Finally, the regression module predicts the pose, the position $\mathbf{p} = \left[ x, y \right]^{T}$ and the orientation $\mathbf{q} = \left[ \cos \theta, \sin \theta \right]^{T}$ from the output of the MHSA module.
It consists of a position branch and an orientation branch as in \cite{Wang2020}, \cite{Wang2022}.
Each branch is composed of consecutive MLPs.
In \cite{Wang2022}, a leaky ReLU activation function was used after each MLP except for the last one in its regression head, but we replace it with the ReLU activation function in our network.
Different from most of the previous studies for end-to-end relocalization, both the position and the orientation are two-dimensional under the assumption that typical serving robots move on planar space.
To take into account the continuity of the rotation angle \cite{Zhou2019}, we present the rotation $\mathbf{q}$ as $\left[ \cos \theta, \sin \theta \right]^{T}$ rather than $\theta$.
To our best knowledge, this work is the first study addressing the end-to-end relocalization for a serving robot based on the camera-2D LiDAR fusion in two-dimensional planar space.

\section{Experiments}
\label{sec:experiments}

\begin{figure}
\begin{center}
    \includegraphics[width=6cm]{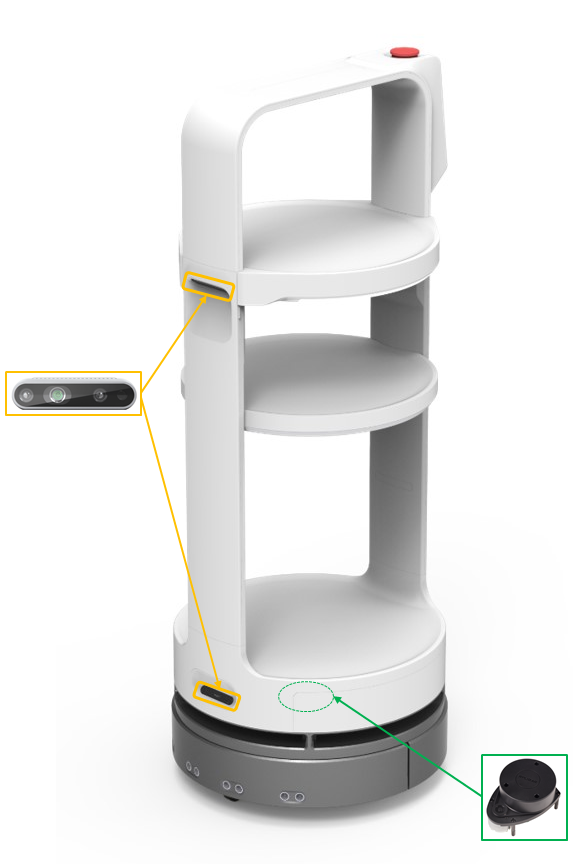}
    \caption{A serving robot with sensors used in this work. The yellow boxes are Intel RealSense D435 cameras and a green box is a SLAMTEC RPLiDAR A1M8 LiDAR. The lower-mounted camera was used for dataset collection together with the LiDAR.}
\label{fig:robot-sensor}
\end{center}
\end{figure}

\begin{figure*}
    \centering
    \subfloat[Set-01]{
        \centering
        \includegraphics[width=0.45\textwidth]{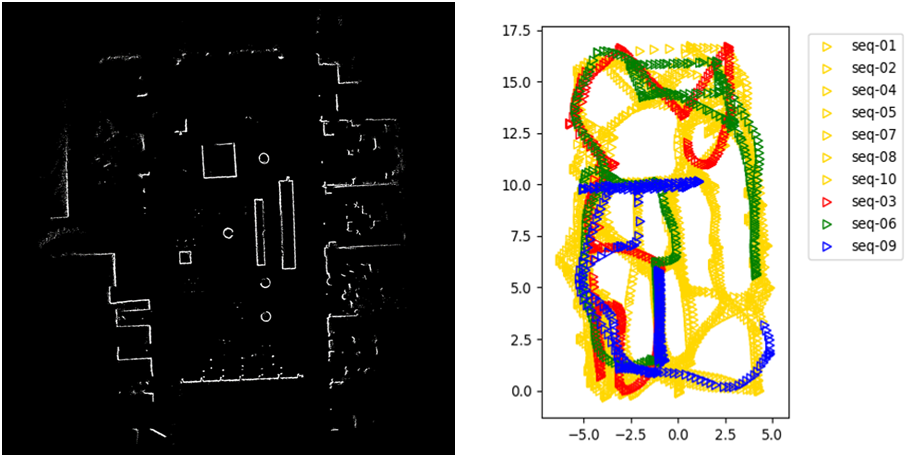}
        \label{fig:small-map}
    }
    \subfloat[Set-02]{
        \centering
        \includegraphics[width=0.45\textwidth]{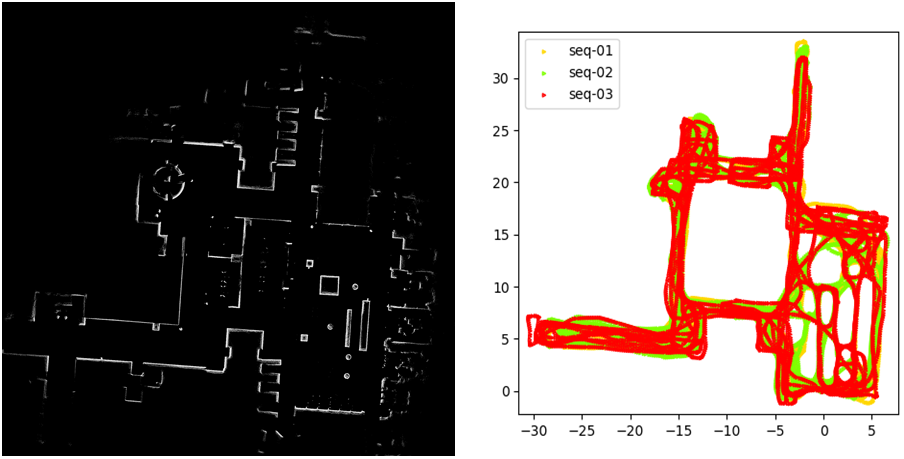}
        \label{fig:large-map}
    }
    \caption{Maps of places where data were collected and the robot trajectories.}
    \label{fig:data-map-traj}
\end{figure*}

\begin{figure*}
    \centering
    \includegraphics[width=17cm]{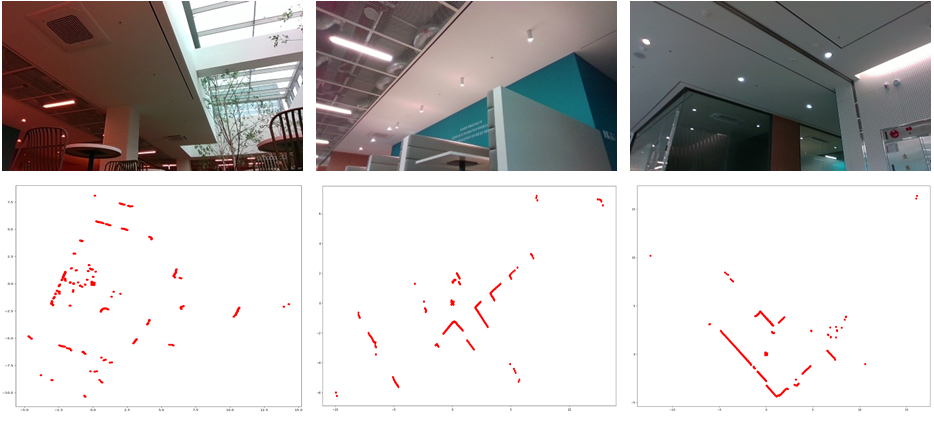}
    \label{fig:data}
    \caption{Example RGB images and 2D point clouds. Each pair of the image and point cloud in the same column are synchronized in time.}
    \label{fig:examples-data}
\end{figure*}

\begin{table}[t]
\caption{Dataset description.}
\label{table:ereon-dataset}
\centering
\setlength{\tabcolsep}{2pt}
\renewcommand{\arraystretch}{1.5}
    \footnotesize
    \begin{tabular}{>{\centering}p{0.18\columnwidth}|>{\centering}p{0.18\columnwidth}|>{\centering}p{0.18\columnwidth}|>{\centering}p{0.18\columnwidth}|>{\centering\arraybackslash}p{0.18\columnwidth}}
        \hline
        Dataset& Sequence& Length& Training& Evaluation \\
        \hline
        \multirow{10}{*}{Set-01}& 
        seq-01& 394& \checkmark&  \\
         & seq-02& 374& \checkmark&  \\
         & seq-03& 389&  & \checkmark \\
         & seq-04& 359& \checkmark&  \\
         & seq-05& 429& \checkmark&  \\
         & seq-06& 401&  & \checkmark \\
         & seq-07& 390& \checkmark&  \\
         & seq-08& 404& \checkmark&  \\
         & seq-09& 408&  & \checkmark \\
         & seq-10& 416& \checkmark&  \\
        \hline
        \multirow{3}{*}{Set-02}& 
        seq-01& 4,820& \checkmark&  \\
         & seq-02& 7,805& \checkmark&  \\
         & seq-03& 6,377&  & \checkmark \\
        \hline
\end{tabular}
\end{table}

\subsection{Dataset}
\label{subsec:dataset}
In order to train and evaluate the neural networks for serving robot relocalization, we constructed a dataset using a commercial serving robot Polaris3D ereon as shown in Fig. \ref{fig:robot-sensor}.
The dataset contains two sets, Set-01 and Set-02.
We gathered the data samples in Set-01 by moving the robot around an area with tables and chairs in our testbed as shown in the left of Fig. \ref{fig:small-map}.
Set-01 originally consisted of a single sequence of 3,964 lengths.
We split it into ten shorter ones as in the right of Fig. \ref{fig:small-map}.
Seven sequences and the others among them were used for training and evaluation, respectively.
In addition, we collected the data samples for Set-02 by operating the robot in a relatively wider area with long corridors as shown in the left of Fig. \ref{fig:large-map}. 
Set-02 consists of three sequences with lengths of 4,820, 7,805, and 6,377 as in the right of Fig. \ref{fig:large-map}.
Two sequences and the other one among them were used for training and evaluation, respectively.
Table \ref{table:ereon-dataset} summarizes the ereon dataset described above.

As shown in Fig. \ref{fig:robot-sensor}, ereon has two cameras, Intel RealSense D435 and one 2D LiDAR, SLAMTEC RPLiDAR A1M8.
The cameras are installed at the side of the upper and lower serving tray, and the LiDAR is mounted in the center of the drive unit located under the lower serving tray.
In order to capture the whole body of people around the robot, the lower and upper cameras face upwards and downwards, respectively, rather than facing straight ahead.
We gathered images and point clouds obtained from the lower camera and the LiDAR because the upper camera was affected by vibration during robot movement.
The obtained RGB images have the size of 420$\times $240 pixels with a frequency of 1.5 Hz.
It may be seen that the image resolution and the saving frequency are low.
The serving robot is equipped with a single-board computer (SBC) instead of a PC or a laptop due to the unit cost of production.
It is not easy in practice to capture and store high-resolution images and 2D LiDAR point clouds at a high frequency on the SBC performing a localization to obtain ground truth pose information mentioned below.
The LiDAR sensor captures 2D point clouds with a range radius of up to 12 meters (m) and a field of view of 360 degrees ($^{\circ}$).
Although it can measure point clouds at a frequency of 8 Hz, we acquired only the point clouds synchronized with images.
Since the angular resolution of the LiDAR sensor is equal to or greater than 0.313$^{\circ}$, the number of 2D points in a point cloud is up to about 1,150.
Fig. \ref{fig:examples-data} shows the example RGB images and 2D point clouds captured by our sensors.
Together with the sensor data, the robot poses corresponding to images and point clouds are necessary to train and evaluate relocalization algorithms.
To do this, we first constructed a map of our testbed using a 2D LiDAR-based SLAM technique and then measured the poses on the constructed map by its localization mode using only the 2D LiDAR.
Since the SBC in ereon has not powerful as mentioned above, 2D LiDAR-based mapping and localization is a better solution than visual SLAM and camera-based localization for real-time navigation.
The localization mode provides a 6-DoF pose with the quaternion representation for orientation.
However, we used the x- and y-axis values for position and the yaw angle value for orientation under the consideration that typical serving robots operate in a flat environment.
Given a quaternion $\left[ q_{x}, q_{y}, q_{z}, q_{w} \right]$, the yaw angle $ \theta $ is calculated as
\begin{equation} \nonumber
    \theta = \arctan{\left( 2\left(q_{x}q_{y} + q_{w}q_{z} \right), 1 - 2\left(q_{y}^{2} + q_{z}^{2} \right) \right).}
\end{equation}

\begin{table}[t]
\caption{Parameters for the set abstraction layers.}
\label{table:set-abst}
\centering
\setlength{\tabcolsep}{2pt}
\renewcommand{\arraystretch}{1.5}
    \footnotesize
    \begin{tabular}{>{\centering}p{0.18\columnwidth}|>{\centering}p{0.18\columnwidth}|>{\centering}p{0.18\columnwidth}|>{\centering}p{0.18\columnwidth}|>{\centering\arraybackslash}p{0.18\columnwidth}}
        \hline
        Layer name& Point num.& Radius& Sample num.& MLP \\
        \hline
        SA1& 256& 0.2& 32& [16, 16, 32] \\
        SA2 & 128& 0.4& 16& [32, 32, 64] \ \\
        SA3 & 64& 0.8& 8& [64, 64, 64] \\ \hline
\end{tabular}
\end{table}

\begin{table}[t]
\caption{Set-01. Average median and mean errors of position and orientation varying the numbers of image and point cloud features $d_{I}$ and $d_{P}$.}
\label{table:dim-output}
\centering
\setlength{\tabcolsep}{2pt}
\renewcommand{\arraystretch}{1.5}
    \footnotesize
    \begin{tabular}{>{\centering}p{0.1\columnwidth}|>{\centering}p{0.2\columnwidth}>{\centering}p{0.2\columnwidth}|>{\centering}p{0.2\columnwidth}>{\centering\arraybackslash}p{0.2\columnwidth}}
        \hline
        $d_{I}$& \multicolumn{2}{c|}{AtLoc (Image)}& \multicolumn{2}{c}{PointLoc (Point cloud)} \\
        \cline{2-5}
        or $d_{P}$ & median& mean& median& mean \\
        \hline
        256& 1.03 m, 22.14$^{\circ}$& 1.48 m, 38.56$^{\circ}$& 1.77 m, 5.77$^{\circ}$ & 2.38 m, 19.44$^{\circ}$ \\
        512& 1.03 m, 23.98$^{\circ}$& 1.49 m, 39.75$^{\circ}$& 1.44 m, 5.03$^{\circ}$& 2.07 m, 17.59$^{\circ}$ \\
        1024& 1.07 m, 24.75$^{\circ}$& 1.50 m, 40.06$^{\circ}$& 1.42 m, 4.10$^{\circ}$& 2.11 m, 16.23$^{\circ}$ \\
        2048& 1.05 m, 25.72$^{\circ}$& 1.52 m, 40.44$^{\circ}$& 1.26 m, 4.20$^{\circ}$& 1.93 m, 16.20$^{\circ}$ \\ \hline
    \end{tabular}
\end{table}

\subsection{Training details}

We implemented and trained our proposed network and other networks presented for the same task under the setting below.
The Adam method was employed as the solver or optimizer. 
The learning rate was set to 0.0001, and the weight decay was determined to be the same value.
The networks were trained up to 1000 epochs using the ereon dataset with a batch size of 256 on a single GPU of NVIDIA GeForce RTX 3090.
As in AtLoc \cite{Wang2020} and PointLoc \cite{Wang2022}, we adopted $\mathcal{L}_{1}$ distances to measure the dissimilarity between the ground truth pose and the estimated pose in our loss function as
\begin{equation} \nonumber
    \lVert \mathbf{p} - \widehat{\mathbf{p}} \rVert_{1} e^{-\beta} + \beta + \lVert \mathbf{q} - \widehat{\mathbf{q}} \rVert_{1} e^{-\gamma} + \gamma,
\label{equ:loss}
\end{equation}
where $\beta$ and $\gamma$ are learnable parameters to balance the position and orientation loss terms.
Their initial values $\beta_{0}$ and $ \gamma_{0}$ were set to $0.0$ and $-3.0$, respectively, as in \cite{Wang2020} and \cite{Wang2022}. 
In the above loss function, we employed the $\mathcal{L}_{1}$ distance between the predicted angle and the ground truth angle instead of the $\mathcal{L}_{1}$ distance between two logarithms of their unit quaternions.
This replacement came from the dimension of the space in which our robot operates.

\begin{table*}[t]
\caption{Set-01. Average median and mean errors of position and orientation using fusion feature with various combinations of $d_{I}$ and $d_{P}$.}
\label{table:Fusion} 
\centering
\setlength{\tabcolsep}{2pt}
\renewcommand{\arraystretch}{1.5}
    \footnotesize
    \begin{tabular}{>{\centering}p{0.05\textwidth}|>{\centering}p{0.1\textwidth}>{\centering}p{0.1\textwidth}|>{\centering}p{0.1\textwidth}>{\centering}p{0.1\textwidth}|>{\centering}p{0.1\textwidth}>{\centering}p{0.1\textwidth}|>{\centering}p{0.1\textwidth}>{\centering\arraybackslash}p{0.1\textwidth}}
        \hline
        \multirow{2}{*}{\backslashbox{$d_{I}$}{$d_{P}$}}& \multicolumn{2}{c|}{256}& \multicolumn{2}{c|}{512}& \multicolumn{2}{c|}{1024}& \multicolumn{2}{c}{2048} \\
        \cline{2-9}
         & median& mean& median& mean& median& mean& median& mean \\ \hline
        256& \cellcolor{yellow!25}0.95 m, 8.19$^{\circ}$& \cellcolor{yellow!25}1.37 m, 34.70$^{\circ}$&  0.96 m, 3.78$^{\circ}$& 1.38 m, 29.95$^{\circ}$& 0.99 m, 3.07$^{\circ}$& 1.39 m, 30.48$^{\circ}$& 0.83 m, 2.38$^{\circ}$& 1.19 m, 22.38$^{\circ}$ \\
        512& 0.97 m, 8.73$^{\circ}$& 1.45 m, 33.72$^{\circ}$& 0.89 m, 4.66$^{\circ}$& 1.34 m, 31.80$^{\circ}$& 0.98 m, 3.77$^{\circ}$& 1.38 m, 29.89$^{\circ}$& 0.87 m, 3.59$^{\circ}$& 1.35 m, 28.64$^{\circ}$ \\
        1024& 0.85 m, 7.81$^{\circ}$& 1.26 m, 32.47$^{\circ}$& 0.86 m, 5.06$^{\circ}$& 1.35 m, 31.98$^{\circ}$& 0.90 m, 3.51$^{\circ}$& 1.32 m, 29.39$^{\circ}$& 0.87 m, 3.18$^{\circ}$& 1.36 m, 27.78$^{\circ}$ \\
        2048& 0.90 m, 5.11$^{\circ}$& 1.34 m, 33.43$^{\circ}$& 0.90 m, 5.83$^{\circ}$& 1.43 m, 32.27$^{\circ}$& 0.85 m, 3.72$^{\circ}$& 1.30 m, 32.56$^{\circ}$& \cellcolor{red!20}0.82 m, 3.51$^{\circ}$& \cellcolor{red!20}1.26 m, 28.22$^{\circ}$ \\
        \hline
    \end{tabular}
\end{table*}

For the purpose of comparison, AtLoc and PointLoc were selected as the image- and the point cloud-based baseline methods for end-to-end serving robot relocalization, respectively.
We also utilized their networks as backbones for image and point cloud feature extraction.
For image feature extraction, as in AtLoc, we scaled the short side of the image to have 256 pixels, then randomly cropped it to 256$\times$256 pixels and normalized the cropped image when training. 
The random cropping was replaced by the center cropping when testing.
For data augmentation, the color jittering method was also conducted by setting the brightness, contrast, and saturation values to 0.7 and the hue value to 0.5. 
The pre-trained ResNet34 with the ImageNet dataset was chosen as the backbone for image feature extraction.
We also applied a dropout operation with a probability of 0.5 whereas no dropout was applied when using BN.
Table \ref{table:set-abst} shows the parameter values of the set abstraction layers in the point cloud feature extractor used in our experiments.
We reduced the numbers of layers and neurons in MLP compared to the original PointLoc setting because the 2D point cloud has fewer input points than the 3D point cloud.
We also employed ReLU instead of LeakyReLU as the activation function in the fully connected layers.
After the feature extractions mentioned above, the image and the point cloud features were concatenated into the fusion feature.
Then, the fusion feature was fed into the several MHSA blocks described in the previous section and the final regression head to provide the pose estimate.
In these procedures, both the image and the point cloud features could be set to 256-, 512-, 1024-, and 2048-dimensional giving 16 combinations of fusion.
Also, we conducted experiments applying MHSA with 1, 2, 4, and 8 heads and 1, 2, 4, and 6 layers.

\begin{figure*}
    \centering
    \subfloat[AtLoc]{
        \centering
        \includegraphics[width=10cm]{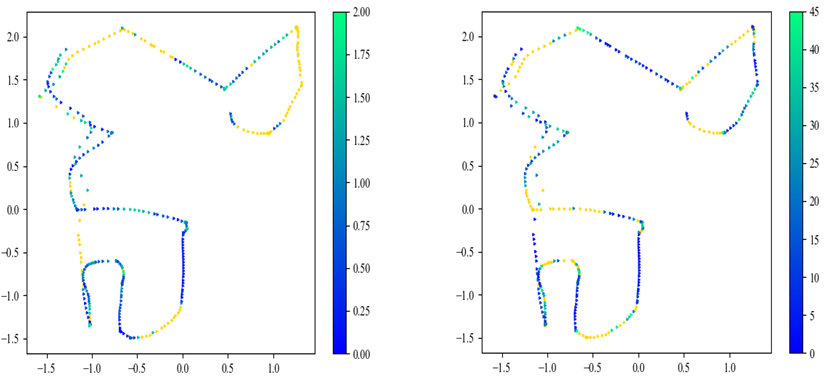}
        \label{fig:errors-atloc-set-01}
    }
    \hfil
    \subfloat[PointLoc]{
        \centering
        \includegraphics[width=10cm]{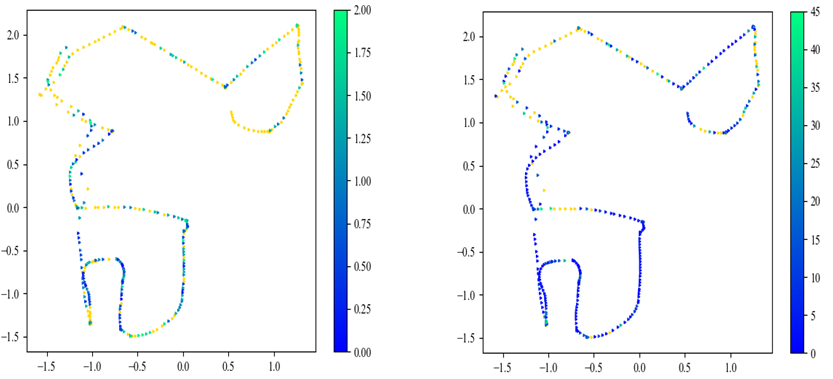}
        \label{fig:errors-pointloc-set-01}
    }
    \hfil
    \subfloat[Concatenation]{
        \centering
        \includegraphics[width=10cm]{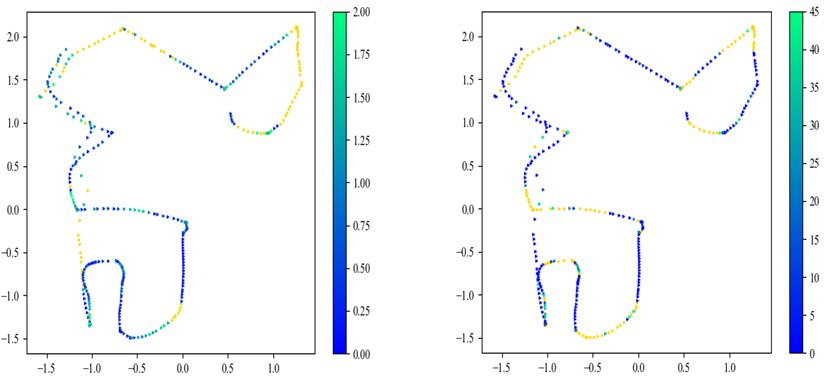}
        \label{fig:errors-fusion-set-01}
    }
    \hfil
    \subfloat[FusionLoc]{
        \centering
        \includegraphics[width=10cm]{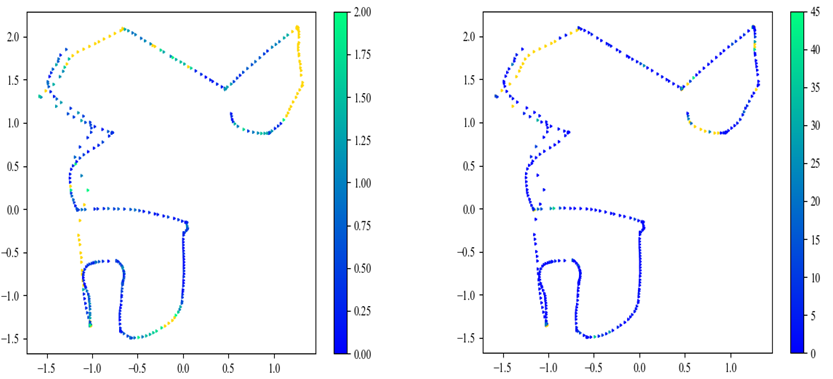}
        \label{fig:errors-fusionloc-set-01}
    }
    \caption{Set-01. Visualization of position (1st column) and orientation (2nd column) errors measured on the evaluation sequence seq-03 when $d_{I}=d_{P}=256$, $N_{h}=2$, and $N_{l}=6$.}
    \label{fig:result-errors-set-01}
\end{figure*}

\subsection{Evaluation results on Set-01} \label{ssec:small-eval}

\textbf{Baselines.} In the original AtLoc and PointLoc, the image and point cloud features were 2048- and 1024-dimensional, respectively.
We first tried to determine the dimensions of the image and point cloud features represented as $d_{I}$ and $d_{P}$, respectively.
To do this, we trained the two methods using the data in the seven training sequences in Set-01 in Table \ref{table:ereon-dataset}.
Then, we measured the median and mean of the position (m) and orientation errors ($^{\circ}$) using the data in each evaluation sequence in the same set for each value of the $d_{I}$ and $d_{P}$.
Most previous end-to-end relocalization studies reported only the median errors.
However, we report the mean errors with the median errors since the latter can not reflect a few large errors.
We computed the averages of the median and mean errors, and Table \ref{table:dim-output} shows the average median and mean errors of position and orientation.
We can see from the table that AtLoc gives lower position errors than PointLoc whereas PointLoc yields lower orientation errors than AtLoc.
Another interesting point is that the tendency between the error values and their feature dimension is opposite, i.e., the error values of AtLoc and PointLoc decrease and increase as their dimension increases, respectively.
Also, in the case of position error, AtLoc provided similar error values varying the dimension of the image features, and the difference was up to 0.04 m and 3.58$^{\circ}$ for position and orientation, respectively.
However, the error values of PointLoc had a large difference, the maximum of which was 0.51 m and 3.24$^{\circ}$.

\noindent
\textbf{A simple fusion.} We present experiments with the fusion feature corresponding to the concatenation of the image and point cloud features extracted by AtLoc and PointLoc without MHSA.
Table \ref{table:Fusion} shows the average median and mean errors of position and orientation depending on the fusion combinations, which have different dimensions of the image and point cloud features.
Note that the position error decreased to less than 1 m by fusing the image and point cloud features in terms of the average median error.
The average median orientation error also maximally decreased to 2.38$^{\circ}$, which was less than the PointLoc's one.
This result clearly shows the benefit of fusing different sensor data.
However, the fusion feature still provided a large average mean error of orientation compared to the point cloud features.
Additionally, we analyzed these results by visualizing position and orientation errors measured using seq-03 as colors.
Fig. \ref{fig:result-errors-set-01} shows the position and orientation errors represented as colors.
In the figure, the positions of the points indicate the ground truth positions.
Also, the yellow indicates outlier values exceeding 2 m in position error and 45$^{\circ}$ in orientation error.
Comparing Fig. \ref{fig:errors-fusion-set-01} to Fig. \ref{fig:errors-atloc-set-01}, we can find the points at which orientation error decreased.
However, we can also see that some points, e.g., on the top right and the bottom middle of Fig. \ref{fig:errors-fusion-set-01}, still have somewhat large orientation errors compared to the corresponding points in Fig. \ref{fig:errors-pointloc-set-01}.

\begin{figure*}
    \centering
    \subfloat[seq-01]{
        \centering
        \includegraphics[width=0.45\textwidth]{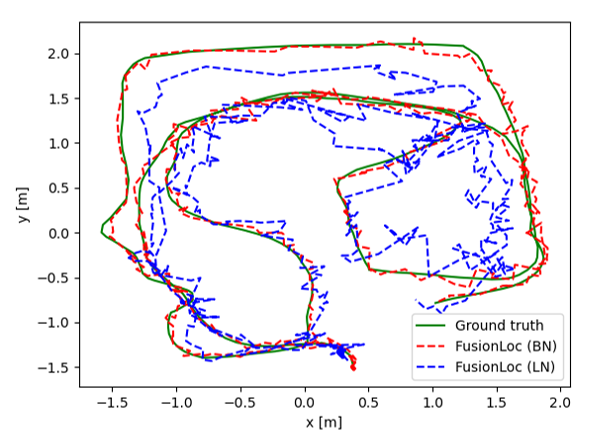}
        \label{fig:seq-01}
    }
    \subfloat[seq-10]{
        \centering
        \includegraphics[width=0.45\textwidth]{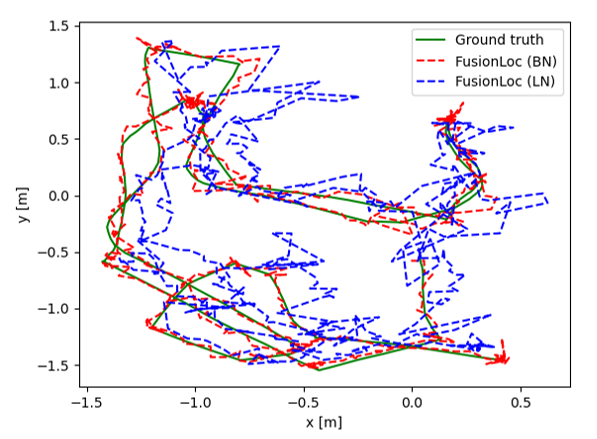}
        \label{fig:seq-10}
    }
    \caption{Set-01. Position prediction comparison using BN and LN in the MHSA module when $d_{I}=d_{P}=256$ and $N_{h}=N_{l}=1$.}
    \label{fig:bn}
\end{figure*}

\begin{table}[t]
\caption{Set-01. Average median and mean errors of position and orientation varying the numbers of attention heads $N_{h}$ and the number of MHSA layers $N_{l}$ in the MHSA module.}
\label{table:set01-MHSA}
\centering
\setlength{\tabcolsep}{2pt}
\renewcommand{\arraystretch}{1.5}
    \footnotesize
    \begin{tabular}{>{\centering}p{0.05\columnwidth}>{\centering}p{0.05\columnwidth}|>{\centering}p{0.2\columnwidth}>{\centering}p{0.2\columnwidth}|>{\centering}p{0.2\columnwidth}>{\centering\arraybackslash}p{0.2\columnwidth}}
        \hline
        \multicolumn{2}{c|}{FusionLoc}& \multicolumn{2}{c|}{($d_{I}$, $d_{P}$)=(256, 256)}& \multicolumn{2}{c}{($d_{I}$, $d_{P}$)=(2048, 2048)} \\
        \hline
        $N_{h}$& $N_{l}$& median& mean& median& mean \\ \hline
        \multirow{4}{*}{1}& 1& 0.88 m, 5.23$^{\circ}$& 1.34 m, 27.06$^{\circ}$
        & 0.65 m, 2.10$^{\circ}$& 1.09 m, \hspace{3pt}8.47$^{\circ}$ \\
         & 2& 0.84 m, 5.22$^{\circ}$& 1.18 m, 24.01$^{\circ}$
         & 0.70 m, 2.15$^{\circ}$& 1.25 m, \hspace{3pt}7.80$^{\circ}$ \\
         & 4& 0.78 m, 2.61$^{\circ}$& 1.19 m, 20.38$^{\circ}$
         & 0.62 m, 2.05$^{\circ}$& 1.13 m, 10.34$^{\circ}$ \\
         & 6& 0.77 m, 2.34$^{\circ}$& 1.14 m, 17.10$^{\circ}$
         & 0.66 m, 1.84$^{\circ}$& 1.00 m, \hspace{3pt}8.30$^{\circ}$ \\
        \hline
        \multirow{4}{*}{2}& 1& 0.91 m, 5.06$^{\circ}$& 1.36 m, 28.16$^{\circ}$
        & 0.61 m, 2.29$^{\circ}$& 1.37 m, \hspace{2pt}8.66$^{\circ}$ \\
         & 2& 0.81 m, 3.15$^{\circ}$& 1.21 m, 23.27$^{\circ}$
         & 0.66 m, 1.96$^{\circ}$& 1.35 m, \hspace{2pt}6.86$^{\circ}$ \\
         & 4& 0.79 m, 2.41$^{\circ}$& 1.14 m, 17.69$^{\circ}$
         & 0.73 m, 1.74$^{\circ}$& 1.25 m, \hspace{2pt}7.75$^{\circ}$ \\
         & 6& \cellcolor{yellow!25}0.72 m, 1.95$^{\circ}$& 1.09 m, 14.19$^{\circ}$
         & 0.65 m, 1.83$^{\circ}$& 1.13 m, \hspace{2pt}9.87$^{\circ}$ \\
        \hline
        \multirow{4}{*}{4}& 1& 0.77 m, 4.74$^{\circ}$& 1.12 m, 25.01$^{\circ}$
        & 0.58 m, 1.68$^{\circ}$& 1.13 m, \hspace{3pt}8.57$^{\circ}$ \\
         & 2& 0.80 m, 3.60$^{\circ}$& 1.19 m, 23.22$^{\circ}$
         & 0.66 m, 2.24$^{\circ}$& 1.28 m, 11.38$^{\circ}$ \\
         & 4& 0.77 m, 1.93$^{\circ}$& 1.20 m, 15.30$^{\circ}$
         & 0.68 m, 2.03$^{\circ}$& 1.21 m, \hspace{3pt}9.25$^{\circ}$ \\
         & 6& 0.74 m, 1.99$^{\circ}$& 1.15 m, 13.41$^{\circ}$
         & 0.63 m, 1.78$^{\circ}$& 1.16 m, \hspace{3pt}8.05$^{\circ}$ \\
        \hline
        \multirow{4}{*}{8}& 1& 0.77 m, 4.89$^{\circ}$& 1.22 m, 26.37$^{\circ}$
        & \cellcolor{red!20}0.58 m, 1.52$^{\circ}$& \cellcolor{red!20}0.99 m, \hspace{3pt}7.25$^{\circ}$ \\
         & 2& 0.83 m, 2.82$^{\circ}$& 1.24 m, 20.69$^{\circ}$
         & 0.63 m, 1.89$^{\circ}$& 1.20 m, \hspace{3pt}6.63$^{\circ}$ \\
         & 4& 0.81 m, 2.33$^{\circ}$& 1.20 m, 19.26$^{\circ}$
         & 0.60 m, 1.96$^{\circ}$& 1.09 m, \hspace{3pt}6.89$^{\circ}$ \\
         & 6& 0.71 m, 2.62$^{\circ}$& \cellcolor{yellow!25}1.07 m, 13.16$^{\circ}$
         & 0.72 m, 2.25$^{\circ}$& 1.28 m, \hspace{3pt}9.56$^{\circ}$ \\
         \hline
    \end{tabular}
\end{table}

\noindent
\textbf{FusionLoc.} In addition to AtLoc, PointLoc, and the simple concatenation, we finally conducted relocalization experiments by adopting repetitive MHSAs taking the fusion feature as input.
As mentioned in the previous section, we applied BN instead of LN to each MHSA block based on the fact that the range of the image feature values was different from that of the point cloud feature values.
Moreover, it was mentioned in \cite{Ba2016} that LN is not as effective in CNN as in recurrent networks.
Fig. \ref{fig:bn} shows the trajectories of the estimated positions by FusionLoc with LN and BN using the training sequences seq-01 and seq-10, respectively.
This result indicates that BN is more effective than LN to train the proposed network for relocalization. 
Table \ref{table:set01-MHSA} shows the results of some ablation studies on the number of attention heads $N_{h}$ and the number of MHSA layers $N_{l}$ in the MSHA module. 
For efficiency, the position and orientation errors were measured under only the settings of ($d_{I}$, $d_{P}$)=(256, 256) and ($d_{I}$, $d_{P}$)=(2048, 2048).
When ($d_{I}$, $d_{P}$)=(2048, 2048), we set the batch size to 64 due to memory limitation.
We can find from the table that the errors generally decrease as each of $N_{l}$ and $N_{h}$ increases in the case of ($d_{I}$, $d_{P}$)=(256, 256).
Especially, most average mean orientation errors become smaller than those of PointLoc when the number of layers is equal to or greater than 4.
This result could not be obtained without the MHSA module as shown in Table \ref{table:Fusion}.
We can also see that the position and orientation errors are minimized when $N_{h}$=2 and $N_{l}$=6 in terms of the average median and when $N_{h}$=8 and $N_{l}$=6 in terms of the average mean in the case of ($d_{I}$, $d_{P}$)=(256, 256).
Compared to the ($d_{I}$, $d_{P}$)=(256, 256) without the MHSA module in Table \ref{table:Fusion}, the minimum error values get smaller by (0.23 m, 6.24$^{\circ}$) and (0.3 m, 21.54$^{\circ}$), respectively.
The improvement can also be found by comparing Fig. \ref{fig:errors-fusion-set-01} and Fig. \ref{fig:errors-fusionloc-set-01}.
And, they are lower than those of ($d_{I}$, $d_{P}$)=(2048, 2048) in Table \ref{table:Fusion} though using the lower numbers of image and point cloud features.
Moreover, the minimum errors decreased more as the MHSA module was applied to the case of ($d_{I}$, $d_{P}$)=(2048, 2048).
Using the MHSA module made differences by (0.24 m, 1.99$^{\circ}$) in terms of average median error and (0.27 m, 20.97$^{\circ}$) in terms of average mean error when ($d_{I}$, $d_{P}$)=(2048, 2048).

\begin{figure*}
    \centering
    \includegraphics[width=16cm]{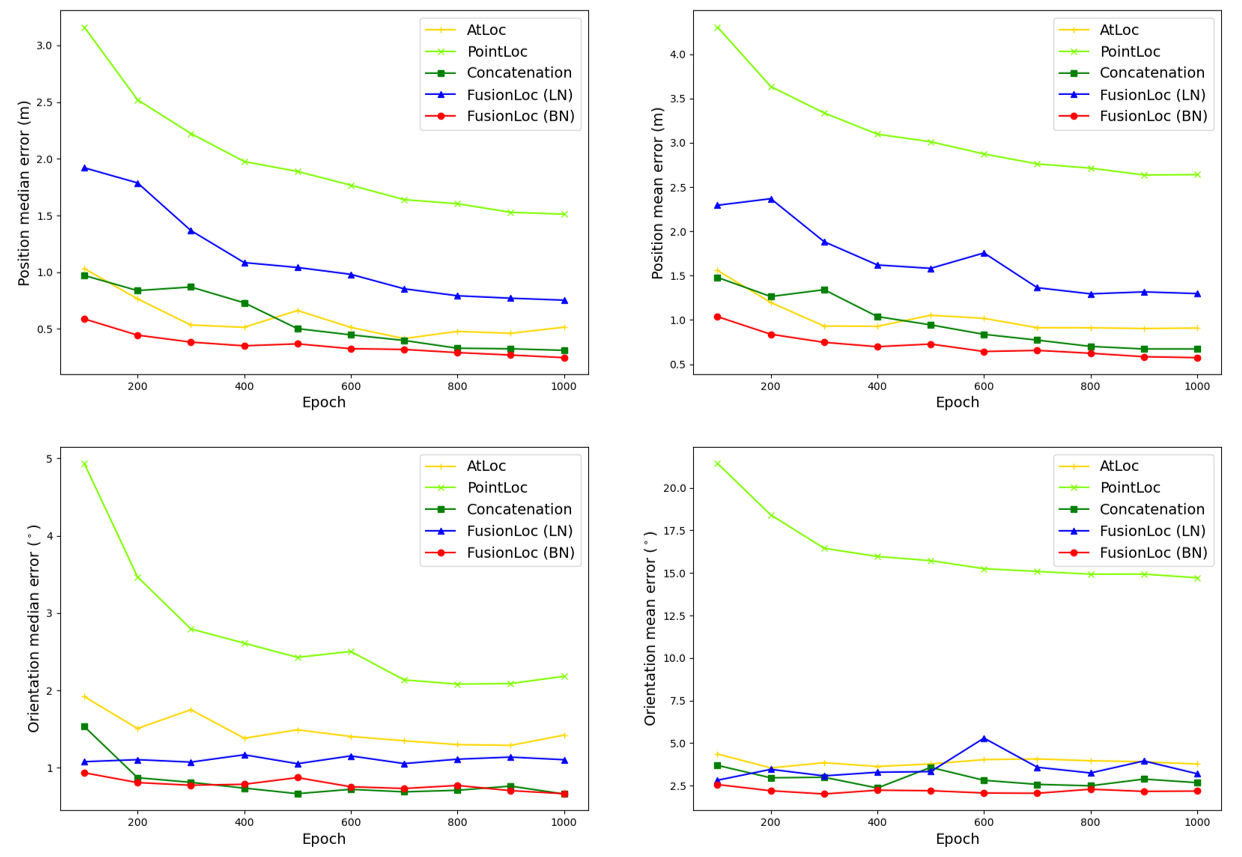}
    \caption{Set-02. Median and mean errors of position and orientation when $d_{I} = d_{P}=256$ and $N_{h}=N_{l}=2$.}
    \label{fig:set02-MHSA}
\end{figure*}

\begin{figure*}
    \centering
    \subfloat[AtLoc]{
        \centering
        \includegraphics[width=12cm]{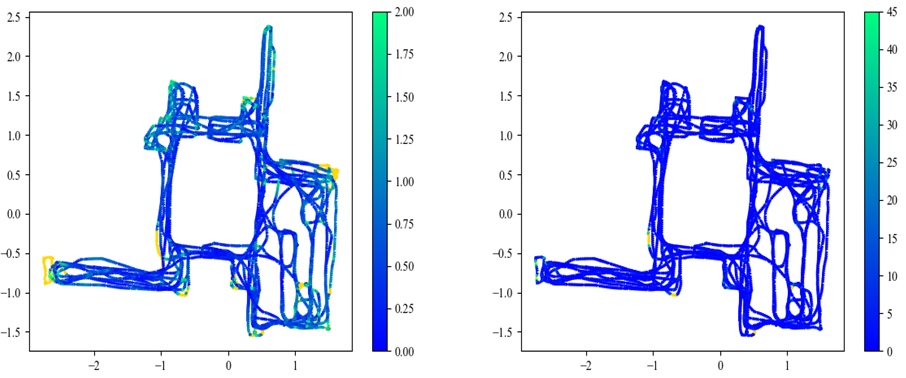}
        \label{fig:errors-atloc-set-02}
    }
    \hfil
    \subfloat[PointLoc]{
        \centering
        \includegraphics[width=12cm]{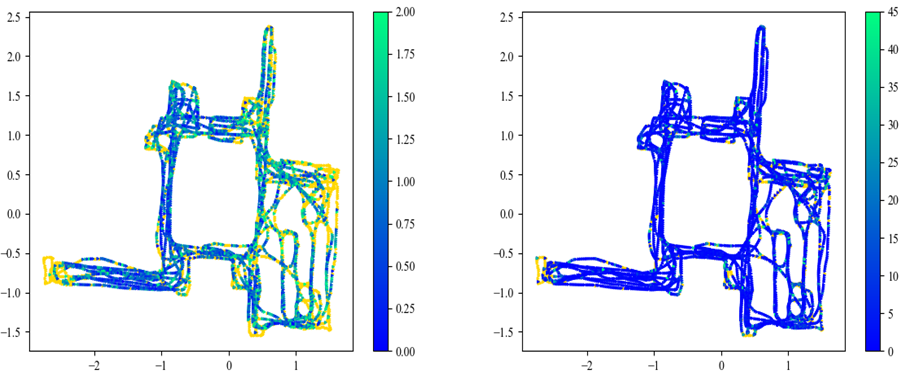}
        \label{fig:errors-pointloc-set-02}
    }
    \hfil
    \subfloat[Concatenation]{
        \centering
        \includegraphics[width=12cm]{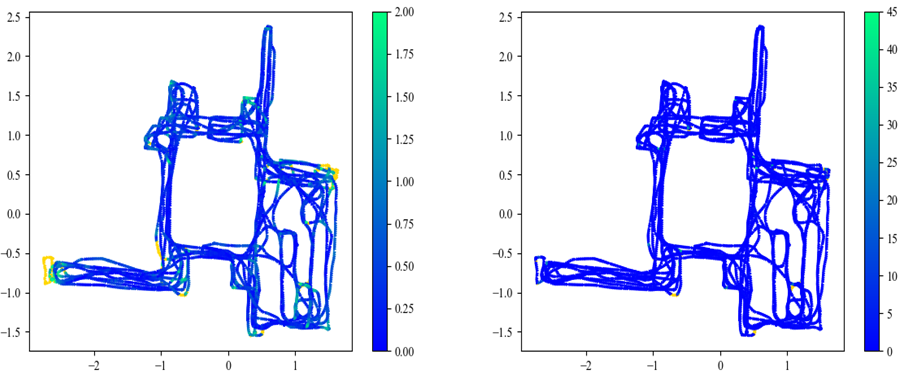}
        \label{fig:errors-fusion-set-02}
    }
    \hfil
    \subfloat[FusionLoc]{
        \centering
        \includegraphics[width=12cm]{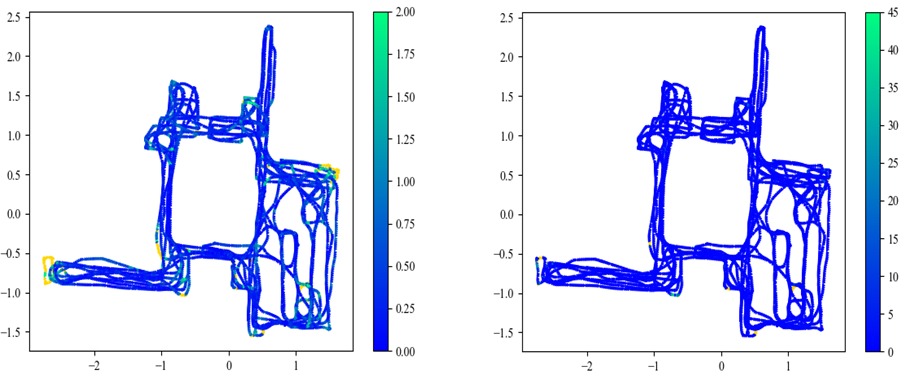}
        \label{fig:errors-fusionloc-set-02}
    }
    \caption{Set-02. Visualization of position (1st column) and orientation (2nd column) errors measured on the evaluation sequence seq-03 when $d_{I}=d_{P}=256$ and $N_{h}=N_{l}=2$.}
    \label{fig:result-errors-set-02}
\end{figure*}

\subsection{Evaluation results on Set-02}
In order to validate the proposed method in a larger place, we compared the position and orientation errors using Set-02 data.
For efficiency, we set to $d_{I}$ = $d_{P}$ = 256.
As in the above experiments using the data in Set-01, we trained every neural network aforementioned using the training sequences in Set-02, and we measured the median and mean errors of position and orientation using the evaluation sequence.
Fig. \ref{fig:set02-MHSA} shows the median and mean errors of AtLoc, PointLoc, the simple concatenation method, FusionLoc using LN, and FusionLoc using BN, which were computed at every 100 up to 1000 epochs.
Different from the results of Set-01 in Table \ref{table:dim-output} that AtLoc was better than PointLoc in terms of the position error whereas PointLoc was better than AtLoc in terms of the orientation error, we can see in Fig. \ref{fig:set02-MHSA} that AtLoc provided lower position and orientation errors than PointLoc. 
This inconsistency may come from the fact that i) Set-02 was collected by making the robot move similar paths multiple times unlike Set-01 as shown in Fig. \ref{fig:data-map-traj}, and ii) the maximum range of the used LiDAR sensor is relatively lower than the area of the place where Set-02 was gathered.
Note that using only the fusion feature could decrease the position and the orientation errors if the learning progresses enough.
Actually, it yielded a more accurate result than AtLoc by 0.24 m and 1.09$^{\circ}$ in terms of the mean errors at 1000 epochs.
We can also see that FusionLoc using LN provided higher errors than the simple concatenation method.
However, FusionLoc using BN gave the minimum errors in position and orientation, which were lower than the concatenation method by 0.1 m and 0.5$^{\circ}$ in terms of the mean error at 1000 epochs.
It corresponds to 15\% and 18\% reductions in the mean position and orientation errors.
Fig. \ref{fig:result-errors-set-02} shows a visualization of the position and orientation errors represented as color using Set-02.
We also figure out from the figure that FusionLoc is more effective than the other methods in the relocalization task.
In summary, our experimental results demonstrate that MHSA can be an effective solution to fuse the features captured by different sensors, and BN is more appropriate than LN in MHSA for robot relocalization based on the camera-2D LiDAR fusion.

\section{Conclusions and Future Works}
\label{sec:conclusions}
In this paper, we proposed FusionLoc, an end-to-end relocalization method for serving robots based on the fusion of RGB images and 2D LiDAR point clouds.
The proposed network performs the pose regression through AtLoc and PointLoc feature extractors, the MHSA module, and the regression module.
To evaluate the proposed network, we constructed a dataset by collecting images, 2D point clouds, and robot poses using a commercial serving robot.
Conducting relocalization experiments using the dataset, FusionLoc showed better performances than the previous relocalization approaches taking only an image or a 2D point cloud as their input.
We observed from the experiments that images and point clouds play a role in complementing the lack of information in each modality.
In particular, MHSA was an effective way to make the interaction between different information contained in the image and point cloud.
Our fusion method using MHSA can help the serving robot find its current pose with less error when it lost its location based on conventional methods such as adaptive Monte Carlo localization.

From the perspective of dataset, our dataset used in the experiments includes only static scenarios at a single site although the proposed method was demonstrated to be more effective than the previous methods.
In future works, we are supposed to supplement the dataset by collecting more data under various variations on dynamic scenarios at several places where the serving robots will be applied.
Using the newly collected dataset, we will also try to additionally reduce relocalization errors with vision Transformers which have shown successes for many computer vision tasks.

\section*{Acknowledgments}
\noindent This work was mainly supported by Electronic and Telecommunications Research Institute (ETRI) grant funded by the Korean government [23ZD1130, Development of ICT Convergence Technology for Daegu-Gyeongbuk Regional Industry]. The first author (J. Lee) was supported by the Industrial Strategic Technology Development Program (20009396) funded By the Ministry of Trade, Industry \& Energy (MOTIE, Korea).

\bibliographystyle{IEEEtran}
\bibliography{relocalization}

\begin{IEEEbiography}[{\includegraphics[width=1in,height=1.25in,clip,keepaspectratio]{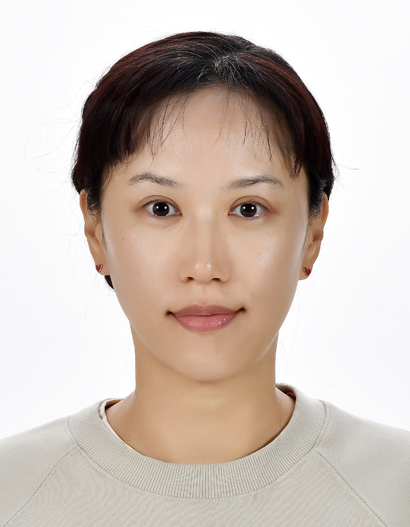}}]{Jieun Lee} received the B.E., M.E., and Ph.D degrees in department of electrical and computer engineering from Ajou University, Korea, in 2009, 2011, and 2019, respectively.

From September to December in 2019, she was a Researcher in Advanced Institute of Convergence Technology, Korea. Since October 2021, she has been a Post Doctoral Researcher in Electronics and Telecommunications Research Institute (ETRI), Korea. Her research interests include computer vision, machine learning, robot perception, and their applications.
\end{IEEEbiography}

\begin{IEEEbiography}[{\includegraphics[width=1in,height=1.25in,clip,keepaspectratio]{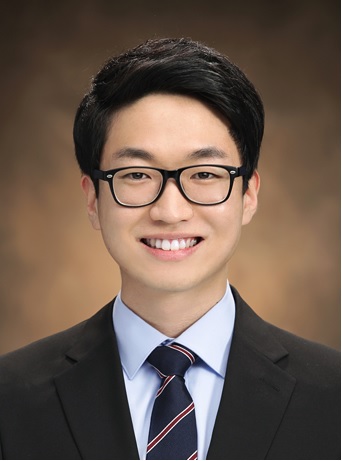}}]{Hakjun Lee} received the B.S. degree in electrical engineering from Chungbuk National University, Cheongju, South Korea and the M.S. and Ph.D. degrees in electrical engineering from the Pohang University of Science and Technology (POSTECH), Pohang, South Korea, in 2014, 2016, and 2020, respectively.

He was a Post Doctoral Researcher in POSTECH from Sep. 2020 to Apr. 2021. He is currently a Senior Researcher with Polaris3D Company, Ltd., Pohang, South Korea. His research interests include service robot, robust control, and navigation system.
\end{IEEEbiography}

\begin{IEEEbiography}[{\includegraphics[width=1in,height=1.25in,clip,keepaspectratio]{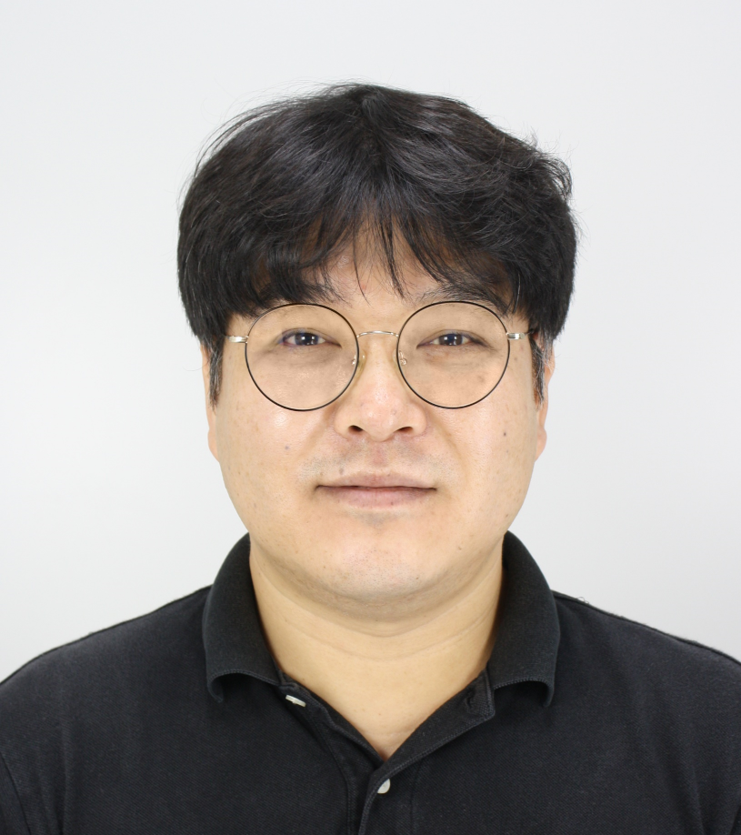}}]{Jiyong Oh} (M'08) received the B.S. degree from the School of Electronic Engineering, Ajou University, Korea in 2004 and the M.S. and Ph.D. degrees from the School of Electrical Engineering and Computer Science, Seoul National University, Korea in 2006 and 2012, respectively.

He was a Post Doctoral Researcher in Sungkyunkwan and Ajou University, Korea in 2012 and 2013, respectively.
From Sept. 2013 (March 2015) to March 2015 (May 2016), he was a Research Fellow (BK Assistant Professor) in the Graduate School of Convergence Science and Technology, Seoul National University, Korea.
Since June 2016, he has been a Senior Researcher in Electronics and Telecommunications Research Institute (ETRI), Korea.
His research interests include computer vision, machine learning, robot perception, and their applications.
\end{IEEEbiography}

\end{document}